\documentclass[letterpaper, 10 pt, conference]{ieeeconf}
\IEEEoverridecommandlockouts
\usepackage{cite}
\usepackage{amsmath,amssymb,amsfonts,bm}
\usepackage{graphicx,hhline,ragged2e}
\usepackage{algorithm}
\usepackage{algcompatible}
\usepackage{booktabs}
\usepackage{tabularx}
\usepackage{textcomp,hyperref,adjustbox}%\usepackage{textcomp,adjustbox} 

\hypersetup{
     colorlinks=true,
     linkcolor=blue,
     filecolor=magenta,
     citecolor = black,      
     urlcolor=black,
     }
\usepackage{xcolor}
\usepackage{array}

\def\BibTeX{{\rm B\kern-.05em{\sc i\kern-.025em b}\kern-.08em
    T\kern-.1667em\lower.7ex\hbox{E}\kern-.125emX}}
    
\title{\LARGE \bf
Behavior Self-Organization Supports Task Inference for Continual Robot Learning
}

\begin{document}

\author{Muhammad Burhan Hafez$^{1}$ and Stefan Wermter$^{1}$%
\thanks{$^{1}$Muhammad Burhan Hafez and Stefan Wermter are with Knowledge Technology, Department of Informatics, University of Hamburg, Germany. 
{\tt\small\{hafez, wermter\}@informatik.uni-hamburg.de}}}

\maketitle
\begin{abstract}
Recent advances in robot learning have enabled robots to become increasingly better at mastering a predefined set of tasks. On the other hand, as humans, we have the ability to learn a growing set of tasks over our lifetime. Continual robot learning is an emerging research direction with the goal of endowing robots with this ability. In order to learn new tasks over time, the robot first needs to infer the task at hand. Task inference, however, has received little attention in the multi-task learning literature. In this paper, we propose a novel approach to continual learning of robotic control tasks. Our approach performs unsupervised learning of behavior embeddings by incrementally self-organizing demonstrated behaviors. Task inference is made by finding the nearest behavior embedding to a demonstrated behavior, which is used together with the environment state as input to a multi-task policy trained with reinforcement learning to optimize performance over tasks. Unlike previous approaches, our approach makes no assumptions about task distribution and requires no task exploration to infer tasks. We evaluate our approach in experiments with concurrently and sequentially presented tasks and show that it outperforms other multi-task learning approaches in terms of generalization performance and convergence speed, particularly in the continual learning setting.
\end{abstract}

\section{Introduction}
 
Recent advances in reinforcement learning (RL) have shown great success, achieving superhuman performance in a variety of challenging board and Atari games \cite{silver2018general, schrittwieser2020mastering} and enabling the acquisition of complex robotic manipulation skills \cite{levine2016end, pinto2016supersizing} through trial and error without any prior knowledge of the task. In RL, however, after a policy for solving a given task has been learned, changing the environment or the task will require training a new policy from scratch. This is especially problematic when learning control tasks on a robot under real-time constraints.\par To tackle this problem and improve the sample efficiency and generalizability on new control tasks, several promising approaches have started to emerge. This includes approaches based on meta-learning \cite{finn2017model, mishra2018simple, 49300}, imitation learning \cite{zhang2018deep, wu2019imitation, balakrishna2020policy}, and transfer learning \cite{peng2018sim, yang2019single}. While these approaches have become essential for multi-task learning, they share a common limitation in that they assume the tasks are known a priori. In other words, a predetermined number of tasks is required by the learning process to define the task space and jointly optimize performance over all tasks. Therefore, they lack the ability to continually learn new control tasks over the lifetime of the agent. Additionally, current approaches to multi-task learning typically require extensive exploration to infer the task before executing the learned policy, particularly in sparse reward environments \cite{houthooft2018evolved, wang2020learning, rakelly2019efficient}. This makes task inference strongly dependent on the exploration policy used to extract task information. Furthermore, such inefficient task exploration is inconsistent with task inference in humans.\par A large body of behavioral and neuroscientific evidence shows that humans can infer a motor task simply by observing a demonstration of the desired behavior, a cognitive function known as goal-directed imitation learning \cite{2000developmentalsci, 2003brainandcognition}. It has been further demonstrated that children encode motor acts they see in terms of goals \cite{2017cognitivescience}. They imitate the action outcomes and not the precise movements. This has challenged the direct-mapping account of imitation, where action observation directly activates the motor system to re-enact the perceived motor behavior, motivated by the mirror neuron system \cite{rizzolatti1996premotor}. A central role for the prefrontal cortex in imitation learning, particularly in the recombination of motor act representations in the mirror neurons to fit the outcome of an observed behavior, was later discovered, supporting the involvement of the mirror system in goal understanding beyond visuomotor mapping \cite{2004neuron, 2014hysiologicalreviews}.\par This imitation learning mechanism is significant for improving higher mental abilities. For instance, inferring task goals from visual demonstrations aids in directing the learner's search for task solutions during motor skill acquisition in physical education \cite{2007sportssciences, 2014experimentalpsy}. Goal-directed imitation is also found to promote cognitive development by enabling the discovery of causal relations and abstract rules through action observation \cite{2010developmentalpsy, 2015frontiersinpsy, 2017experimentalchildpsy}.\par In this paper, inspired by human goal-directed imitation, we propose a novel continual robot learning approach to address the above two limitations of existing robot multi-task learning approaches, namely the assumed prior knowledge about the task space and the inefficient task inference process. Our approach performs unsupervised learning of behavior embeddings and learns a control policy on top of the learned embedding space. Behavior embeddings are learned by incrementally self-organizing latent representations of visual demonstrations of motor behaviors. This allows learning a growing set of behaviors and enables behavior embeddings to capture task-discriminative contextual information. A task can thus be inferred simply by finding the nearest behavior embedding w.r.t. an input demonstration of a motor behavior, which accords with the neural and behavioral findings on task inference in humans. Besides, we use the imitation error in the learned embedding space to derive a behavior-matching intrinsic reward, encouraging the alignment between the demonstrated and policy-generated trajectories.\par Our contributions can be summarized as follows. 1) To the best of our knowledge, our work is the first to explore demonstration-based behavior self-organization for robotic task inference. 2) We propose a behavior-matching intrinsic reward based on the distance in the behavior embedding space between demonstrated and generated trajectories. 3) We empirically evaluate the proposed approach in experiments with concurrently and sequentially presented tasks, comparing it to other multi-task learning approaches.

\section{Related Work}
Multi-task learning approaches can be categorized in terms of how task inference is addressed into the following groups: 

\noindent\textbf{Task-agnostic:} Achille et al. \cite{achille2019task2vec} propose an approach to the meta-learning problem of model selection. The approach computes vector representations of classification tasks. The representation is based on the amount of information each parameter of a model trained on a task contains about that task, measured by the Fisher information matrix of the model, and is therefore independent of task-specific model outputs. This allows for defining a similarity metric between a target task and models trained on different tasks, which subsequently helps in selecting the best pretrained model for a novel task. Although only applied to supervised image classification, the proposed approach is shown to have higher performance and computational efficiency than the brute-force method of finetuning and evaluating all available models. To enable fast learning of new tasks, Finn et al. \cite{finn2017model} propose a few-shot meta-learning algorithm. Given a distribution over tasks, the algorithm trains a model to maximize the improvements on the loss function for a set of sampled tasks resulting from one gradient update to the model's parameters on each task loss. The results show that the proposed approach outperforms other baselines on learning new tasks including finetuning a model pretrained on the same set of tasks. The approach, however, assumes the task distribution is known beforehand and performs expensive Hessian computations in each optimization step. It also requires collecting action trajectories at test time that are used for adapting to new control tasks. In a different direction, Garcia and Thomas \cite{garcia2019meta} decompose the control policy into a task-dependent exploitation policy and a task-independent exploration policy. The former is trained on trajectories of a given task and the latter is trained on trajectories of tasks sampled from a known distribution. The approach uses $\epsilon$-greedy to arbitrate between the two policies at each timestep and is found to improve the learning speed on novel tasks.

\noindent\textbf{Fixed task representation:} 
Rusu et al. \cite{rusu2016policy} propose supervised training of a multi-task policy from trajectory sets each generated by a task-specific expert policy trained through reinforcement learning. The multi-task policy has a separate output layer for each task and uses a task label as input to select which layer to update. While the proposed approach mitigates the learning interference of training a single policy network on a set of tasks, it requires adding a separate output layer and a supervised training set each time a new task is introduced. Similar to \cite{rusu2016policy}, Shao et al. \cite{shao2020concept2robot} train a multi-task policy using supervised learning on expert trajectories from single-task policies. However, the tasks are represented by a task embedding that combines an encoded natural language instruction and image observation and is fed as input to the policies. One limitation to the proposed approach when adding new tasks is that it requires training new policy networks to perform these tasks and retraining the multi-task policy on rollouts from the new policies. To improve the efficiency of learning action policies for multiple tasks, active selection of tasks has been proposed \cite{baranes2013active, santucci2016grail}. In contrast to random or sequential selection, these approaches use the competence improvement in achieving a task as an intrinsic motivation to guide task selection towards tasks where maximum learning progress is expected. Santucci et al. \cite{santucci2019autonomous} extend active task selection to the case of interdependent tasks by proposing a hierarchical reinforcement learning model where active task selection is cast as a Markovian decision process. This line of research, however, does not address the generalization of the learned policies to novel tasks. Moreover, it requires the tasks to be explicitly defined in terms of spatial goals in the environment in order to be used as input to the learning system.

\noindent\textbf{Learned task representation:}
Oh et al. \cite{oh2017zero} assume tasks are defined by a combination of disentangled parameters and learn an embedding over tasks. To enable generalization over unseen configurations of task parameters, they propose using an analogy-making objective that encourages the learned embedding to capture task similarities during policy optimization. While the results show a higher success rate over unseen tasks compared to a baseline without the analogy-making objective, the proposed method requires prior knowledge given by a handcrafted set of analogies between tasks, which does not allow for continually introducing new tasks during training or for task generalization if analogies are unknown or cannot be defined. As opposed to previous approaches to multi-task learning that often ignore task inference, Wang et al. \cite{wang2020learning} propose a variational meta-RL method with a latent task variable whose posterior, parameterized by a task encoder network, is learned jointly with the task-specific action policy. In their method, the authors train a separate exploration policy for collecting informative experiences for task inference by using the information gain from an experience as a reward. During testing, the exploration policy collects experiences sufficient to infer the task posterior which the action policy uses to perform the task, minimizing train-test distribution mismatch. Compared to prior works, the method achieves higher sample efficiency during training and testing on various benchmarks. Nevertheless, it requires task exploration before testing the action policy on a target task, making task performance dependent on the ability of the exploration policy to infer tasks correctly.

\section{Continual Multi-Task Learning from Self-Organized Behaviors}

In this section, we introduce our approach to continual learning of robotic control tasks. Our approach learns, in an unsupervised manner, a behavior embedding by incrementally self-organizing demonstrated behaviors. Task inference is performed by finding the nearest behavior embedding to a demonstrated behavior, which is then used together with the environment state as input to a multi-task policy trained with reinforcement learning to optimize performance over tasks.\par We start with describing the neural architecture used to learn vector representations of demonstrated behaviors. Next, the growing self-organization of the behavior space is explained. This is followed by presenting our algorithm for training a multi-task policy to solve tasks inferred from behavior self-organization.

\subsection{Latent Encoding of Behavior Demonstrations}
\label{sec3a}
\begin{figure}[ht]
	\centering
		\includegraphics[width=0.91\linewidth]{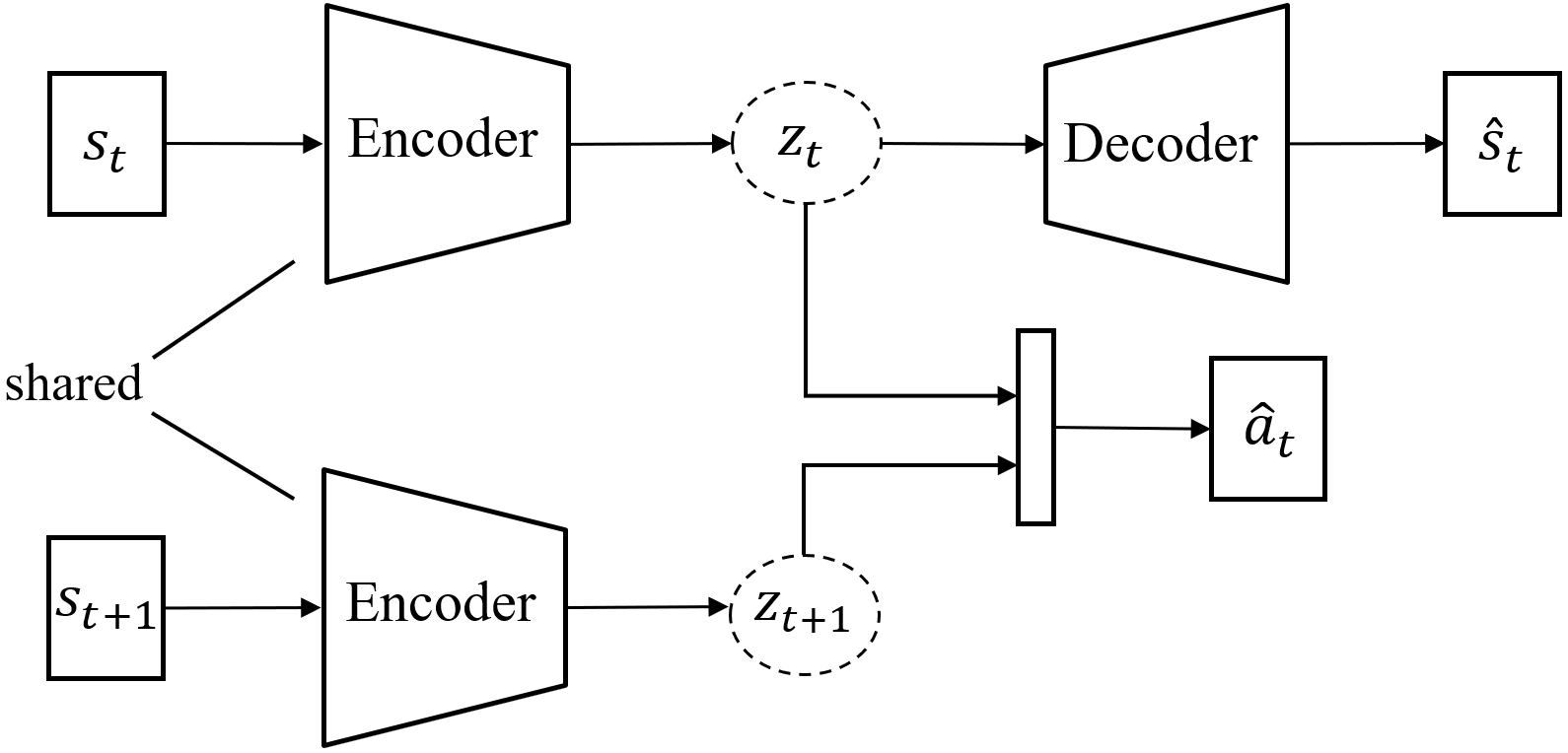}
		\caption{Variational Autoencoder (VAE) model: The input to the model is transition triples of the form $(s_{t},a_{t},s_{t+1})$ in a trajectory. The encoder learns a representation $z_t$ that jointly optimizes the reconstruction $\hat{s}_t$ of the original input $s_t$ from the decoder and the prediction $\hat{a}_t$ of the performed action $a_t$ from an inverse model. The inverse model is approximated by a neural network that takes in the encoder's representation for the states $s_t$ and $s_{t+1}$ and outputs a prediction $\hat{a}_t$ (please see Sec. \ref{sec4} for details on the model architecture and hyperparameters).}
		\label{fig:1}
\end{figure}

The behavior demonstrations we consider here are sequences of visual observations. Specifically, each observation in a demonstrated behavior is a high-dimensional frame from a video sequence. In general, any representation of a video frame, including raw image pixels, can be used in our approach. However, we use a Variational Autoencoder (VAE) \cite{kingma2014auto} to learn a low-dimensional abstract representation of each frame. Using an autoencoder trained only to reconstruct the input at the pixel level can result in learning features that are unlikely to be useful for future tasks. Therefore, we train the autoencoder to jointly optimize the reconstruction of the input and the prediction of an inverse model (i.e., a model that predicts the action taken between two consecutive states). This encourages the VAE to encode information corresponding to changes in the environment that are the result of the robot/demonstrator's own actions and not randomness or noise in the environment (see Fig. \ref{fig:1}). Given an encoder function $\psi$ defined as $\psi(s)=z$, where $s$ is an environment state and $z$ is the feature vector of $s$, an inverse model $\mathcal{M}$ can be defined as $\mathcal{M}(\psi(s_i),\psi(s_{i+1}))=\hat{a}_i$, where $\hat{a}_i$ is the prediction of the action responsible for the transition from $s_i$ to $s_{i+1}$. A feature representation is then learned that jointly optimizes the variational autoencoder loss \cite{kingma2014auto} and the inverse model loss $\frac{1}{2} \Vert \hat{a}_i - a_i \Vert _{2}^{2}$, given a transition $(s_{i},a_{i},s_{i+1})$. The VAE model is trained unsupervised on visual trajectories corresponding to demonstrations of different motor behaviors.

\begin{figure}[t]
	\centering
		\includegraphics[width=\columnwidth, height=4.0cm]{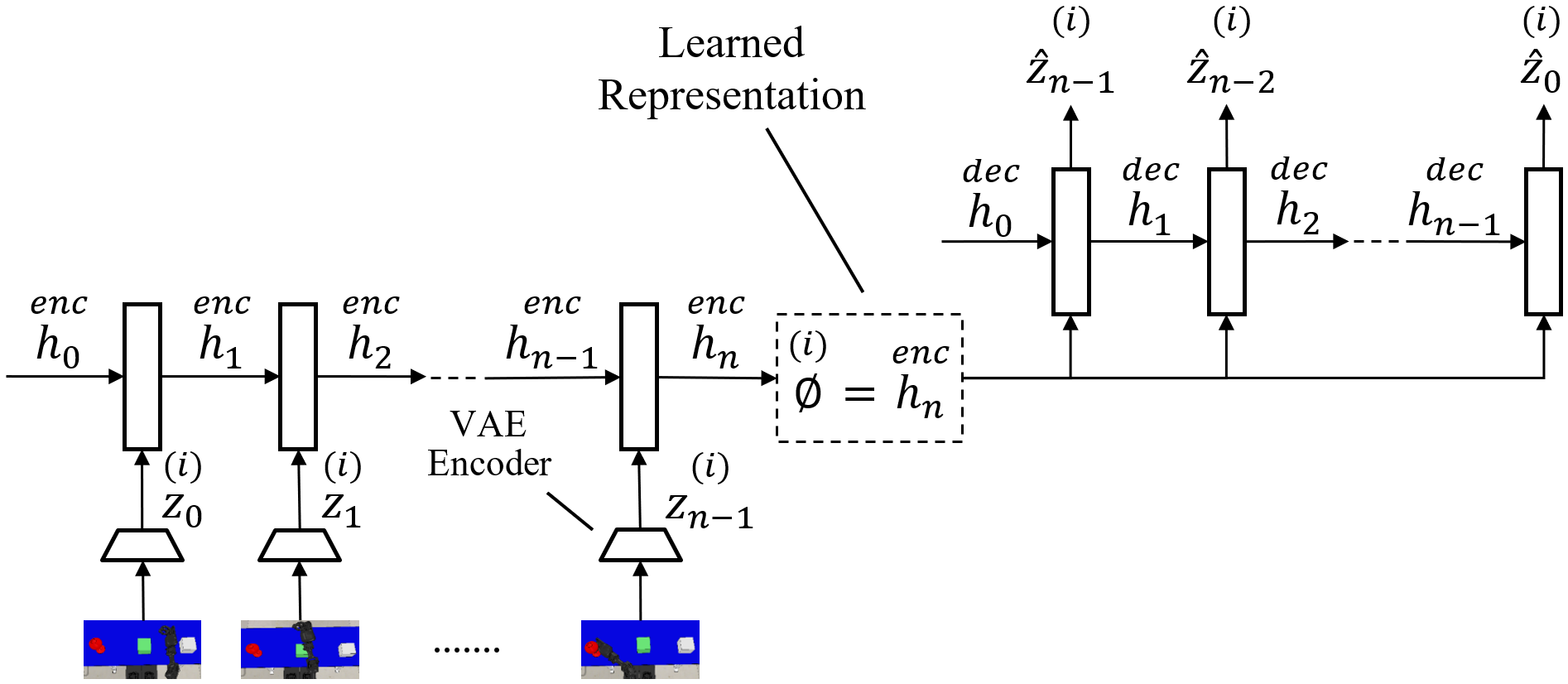}
		\caption{LSTM Autoencoder (LAE) model: The input consists of $N$ feature vector sequences $\{z^{(i)}\}_{i\in 1,...,N}$ generated by the encoder of the trained VAE model. The hidden state $h^{enc}_n$ of the encoder LSTM after the last feature vector $z_{n-1}$ has been read is the latent representation $\phi$ of the entire sequence. Using this representation, the decoder LSTM reconstructs the input sequence in reverse order. The reconstructed sequence and hidden state of the decoder LSTM are denoted by $\hat{z}^{(i)}$ and $h^{dec}$ respectively (please see Sec. \ref{sec4} for details on the model architecture and hyperparameters).}
		\label{fig:2}
\end{figure}

We now aim to learn a representation that can capture the temporal structure of a behavior demonstration. For this reason, we propose to use the LSTM autoencoder (LAE) model shown in Fig. \ref{fig:2}. Long Short-Term Memory (LSTM) \cite{lstm1997long} is a recurrent neural network that has been successfully applied to many domains for processing sequential data. The LAE model consists of an encoder LSTM and a decoder LSTM. The encoder LSTM learns to map an input sequence of feature vectors to a fixed-length vector $\phi$ that contains information about the entire sequence. This vector is fed to the decoder LSTM that learns to reconstruct the input sequence. The LAE model is trained unsupervised on the same trajectories of behavior demonstrations used to train the VAE model. After training is done, the LAE encoder is used to provide a latent representation of input demonstrations which encodes the temporal context of the observed behavior for further processing by our approach.\par

\subsection{Behavior Self-Organization}
\label{sec3b}
Our approach incrementally learns a mapping from an input space of demonstrations to a space where similar behaviors are located close to each other (behavior embeddings). This is performed by building a network that self-organizes the learned latent space of demonstrations. While any growing self-organizing network with no predefined size and structure can be used for this purpose, we choose the Grow When Required (GWR) network \cite{marsland2002self}. This is because the GWR network grows whenever it does not have a sufficiently close match to an input stimulus, as opposed to other criteria, which in our case helps in the addition of novel behaviors once discovered. We call our GWR-based model for unsupervised learning of behavior embeddings GWR-B. \par
\begin{figure}[t]
	\centering
		\includegraphics[width=\linewidth]{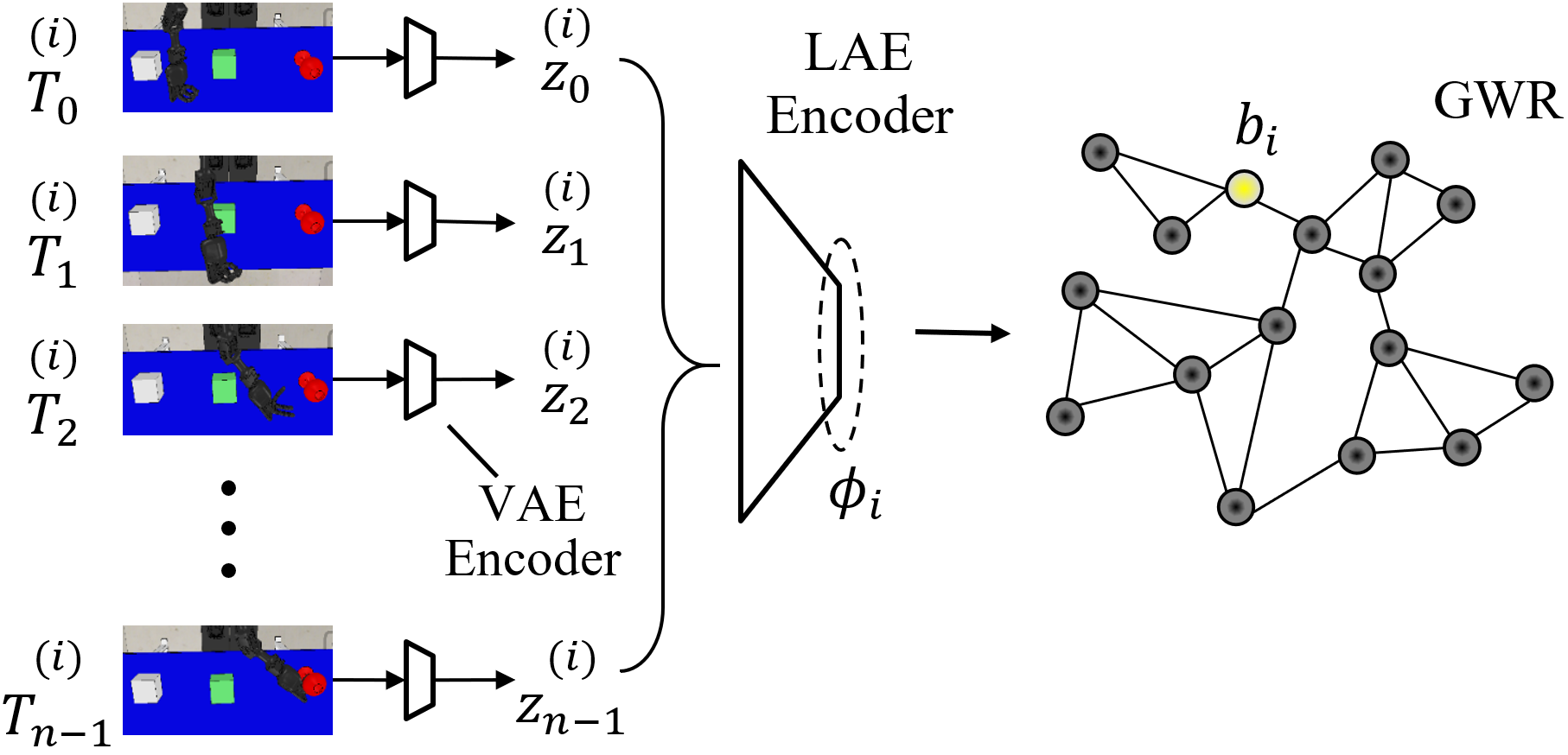}
		\caption{GWR-B model: Given a visual trajectory $T^{(i)}$ of $n$ frames representing a first-person demonstration of a control behavior, the encoder of the VAE model outputs a feature vector $z$ for each frame. The temporal sequence of feature vectors is used as input to the LAE model, whose encoder in turn outputs a latent representation $\phi_i$ of the input demonstration. The GWR network then finds the best-matching node w.r.t. $\phi_i$ (the node in yellow) using Eq. \eqref{eq:eq1} and updates its topology. The behavior embedding $b_i$ of the input demonstration is the weight vector of the best-matching node.}
		\label{fig:3}
\end{figure}

 The growing network in our GWR-B model is defined by a set of nodes $V$, each with a weight vector $w$, and a set of edges connecting each node to its neighbors. At the beginning of learning, the network has two connected nodes with random weights. A learning iteration $i$ starts by sampling an input demonstration $\phi_i$ and finding the best-matching node w.r.t. $\phi_i$:
\begin{equation}
\label{eq:eq1}
c = \arg\min_{j\in V}  \Vert \phi_i - w_j \Vert_{2}.
\end{equation}
An edge is then added between the best- and second-best matching nodes, $c$ and $c'$ respectively, if it does not exist. Otherwise, the age of the edge is set to 0. The activity $a$ of the best-matching node is computed based on the Euclidean distance between its weight vector $w_{c}$ and the input $\phi_i$:
\begin{equation}
\label{eq:eq2}
a = \exp{(-\Vert \phi_i - w_{c} \Vert_{2})}.
\end{equation} 
Storing information about how often a node has fired is done by using a habituation counter that starts at 1 and decreases exponentially to 0 each time that node is the best match. A new node $v$ is added if $a$ falls below activity threshold $a_T$ and the best-matching node's habituation counter $h_{c}$ falls below habituation threshold $h_T$. 
The weight of the new node $w_v$ is set to $(w_{c}+\phi_i)/2$. Two edges are added to connect $v$ with $c$ and $c'$, and the edge between $c$ and $c'$ is removed. If no new node is added, the weights of the best-matching node and its topological neighbors $k$ are adapted towards the input $\phi_i$:
\begin{equation}
\label{eq:eq3}
\Delta w_{c} = \epsilon_{c} \times h_{c} \times (\phi_i - w_{c}),
\end{equation}
\begin{equation}
\label{eq:eq4}
\Delta w_k = \epsilon_n \times h_k \times (\phi_i - w_k),
\end{equation}
where $0<\epsilon_n<\epsilon_{c}<1$ and $h_k$ is the habituation counter for node $k$. Similarly, the habituation counters of the best-matching node and its neighbors are decreased as follows:
\begin{equation}
\label{eq:eq5}
h_c = h_0 - \frac{(1-e^{\frac{-\alpha_ci}{\tau_{c}}})}{\alpha_{c}},
\end{equation}
\begin{equation}
\label{eq:eq6}
h_k = h_0 - \frac{(1-e^{\frac{-\alpha_ni}{\tau_n}})}{\alpha_n},
\end{equation}
where $h_0$ is the initial habituation value and $\alpha$ and $\tau$ are constants controlling the behaviour of the habituation curve. Equations \eqref{eq:eq3}{--}\eqref{eq:eq6} account for the fact that nodes that frequently fire (i.e., $h\ll1$) are considered well-trained and require no further adaptation. Without this habituation mechanism, nodes will continue to move and the network will not converge. Finally, the age of all edges emanating from the best-matching node is incremented and the edges older than a given threshold $\kappa$ are removed. If a node has no remaining edges, that node is removed too. This process is repeated until a maximum number of iterations is reached. An illustration of our GWR-B model is shown in Fig. \ref{fig:3}.

\subsection{Behavior-Guided Policy Optimization}
In order to learn a policy for solving multiple control tasks, the robot first needs to infer the task at hand. A task inference mechanism that is efficient and applicable to an open-ended number of tasks is therefore critical to continual robot learning. Inspired by demonstration-based task inference in humans, we propose an algorithm that infers a task by using the GWR-B model to obtain a behavior embedding that best matches a visual demonstration of the task and trains a control policy over the joint state-behavior space with reinforcement learning. In this way, the algorithm goes beyond task inference in that it infers the intended behavior behind a demonstration, since tasks can often be achieved by following different motor behaviors. The complete algorithm, which we call Behavior-Guided Policy Optimization (BGPO), is given in Algorithm \ref{alg:alg1}.

At the start of each episode, a visual demonstration $d$ of a task is presented to the robot. The LAE encoder computes the corresponding latent representation $\phi_d$. The GWR-B model then finds the best-matching node $c$ w.r.t. $\phi_d$ (see Eq.\eqref{eq:eq1}) and uses the weight vector $w_c$ of node $c$ as the behavior embedding $b$. Actions are generated by a policy $\pi_{\theta}$ that takes as input the state feature vector $z$ and behavior embedding $b$ and is parameterized by policy parameters $\theta$. Besides the extrinsic reward from the environment $r^{ext}$, an intrinsic reward is computed at the end of each episode based on the imitation error defined as the Euclidean distance between $b$ and the latent representation of the generated trajectory $\phi_g$:
\begin{equation}
\label{eq:eq7}
r^{int}=\exp{(- \Vert b - \phi_g \Vert_{2})}.
\end{equation}
This encourages the alignment between demonstrated and generated trajectories in behavior space rather than action space. Since the behavior embedding encodes sufficient information for task inference, the proposed behavior-matching intrinsic reward is in line with the neuroscientific findings on goal-directed imitation in humans, as opposed to imitating the precise actions. The RL algorithm used for training the policy is Deep Deterministic Policy Gradient (DDPG) \cite{lillicrap2016continuous}. DDPG updates the policy by gradient ascent on the action-value ($Q$-)function with a minibatch sampled from the replay buffer $B$:
\begin{equation}
\label{eq:eq8}
\Delta \theta = \frac{\alpha}{n}  \sum _{i}^{}\triangledown _{a}Q \left( s,a  \vert   \varphi \right) \vert _{s= s_{i}, a=  \pi  \left( s_{i} \right) } \triangledown _{ \theta}\pi \left( s \vert \theta \right) \vert _{s= s_{i}},
\end{equation}
where $\alpha$ is the step size, $n$ is the minibatch size, and $Q(s,a)$ is the $Q$-function parameterized by $\varphi$ and gives an estimate of the expected future reward for taking action $a$ in state $s$. In our implementation, the state feature vector $z$ and the behavior embedding $b$ replace $s$ as input to $Q$ and $\pi$.

\begin{figure}[t]
  \centering
  \begin{minipage}{\linewidth}
  \begin{algorithm}[H]
   	\footnotesize
   	\caption{Behavior-Guided Policy Optimization (BGPO)}
   	\label{alg:alg1}
   	\begin{algorithmic}[1]
  \STATE 
  %\\ \hspace{0.87cm} \begin{tabular}{rl} \end {tabular}
  \textbf{Given:} an RL algorithm $f_{RL}$, a VAE encoder  $f_{V}$, an LAE encoder $f_{L}$
  \STATE Initialize policy parameters $\theta$ randomly
  \STATE Initialize replay buffer $B \leftarrow \{\}$ and generated trajectory $g \leftarrow ()$ 
  \STATE Initialize GWR-B model
  \FOR {$episode = 1, E$}
  \STATE Sample demonstration $d = (x_0,...,x_n)$ 
  \STATE $\phi_d \leftarrow f_L(f_V(x_0),...,f_V(x_n))$
  \STATE Find best-matching node $c$ w.r.t. $\phi_d$ using Eq. \eqref{eq:eq1}
  \STATE Compute behavior embedding $b \leftarrow w_c$
  \STATE Adjust the GWR network using Eq. \eqref{eq:eq2}{--}\eqref{eq:eq6}
  \STATE Sample initial state $s$
  \WHILE {not terminal}
  \STATE $z \leftarrow f_V(s)$
  \STATE $g \leftarrow g \oplus (z)$
  \STATE Sample action $a \sim \pi_{\theta}(a \vert z,b)$
  \STATE Execute $a$ and observe $r^{ext}$ and $s'$
  \STATE $z' \leftarrow f_V(s')$
  \IF {$s'$ is terminal}
  \STATE $g \leftarrow g \oplus (z')$
  \STATE $\phi_g \leftarrow f_L(g)$
  \STATE $r^{int}=\exp{(- \Vert b - \phi_g \Vert_{2})}$
  \STATE Store $(z,b,a,r^{ext}+\beta r^{int},z')$ in $B$
  \ELSE 
  \STATE Store $(z,b,a,r^{ext},z')$ in $B$
  \ENDIF
  \STATE Pass $B$ to $f_{RL}$ to update $\theta$
  \STATE $s \leftarrow s'$
  \ENDWHILE
  \STATE $g \leftarrow ()$
  \ENDFOR
  \STATE {\textbf{return} optimized policy $\pi_{\theta}$}
   	\end{algorithmic}
   \end{algorithm} 
     \end{minipage}
\end{figure}

\section{Experimental Results}
\label{sec4}
We evaluate BGPO on continual learning of robotic control tasks and compare it against the following well-known multi-task learning baselines: Model-Agnostic Meta-Learning (MAML) \cite{finn2017model} and Distillation with Negative Log Likelihood loss (Dist-NLL) \cite{rusu2016policy}. \footnote{A video showing the experimental results on the NICO robot is available at https://youtu.be/4B4GL-HPQjg} We further conduct an ablation study to demonstrate the effect of the GWR-B model and the behavior-matching intrinsic reward on the learning performance of BGPO. \par
\subsection{Experimental Setup}

\noindent\textbf{Hyperparameter settings:} 
The encoder for the VAE model has three 3$\times$3 convolutions with 32, 64, and 128 channels respectively. Each convolution is followed by ReLU activation and 2$\times$2 max-pooling. This is followed by two dense layers, each of 64 units, generating the mean $\mu$ and standard deviation $\sigma$ of a multivariate Gaussian $\mathcal{N}(\mu,\sigma^2 I)$ from which the latent vector $z$ is sampled. The decoder is a mirror of the encoder except that the output layer has a sigmoid activation. The inverse model jointly optimizing $z$ is a 1-layer MLP with tanh activation. The LAE model uses 32 LSTM units (for both the encoder and decoder) and an output projection layer of 64 linear units in the decoder. The GWR-B hyperparameters are listed in Table \ref{table:table1}. The policy and $Q$-functions are parameterized by a 2-layer MLP each. The hidden layer is 64-dimensional with ReLU activation. The output layer is 1-dimensional with linear activation in the $Q$-network and 4-dimensional with tanh activation in the policy network. The size of the replay buffer is set to $10^6$. All networks are trained using the Adam optimizer \cite{kingma2014adam} with learning rate 0.001 and batch size 256, except for the VAE networks that use a learning rate of 0.0001. The intrinsic reward weighting coefficient $\beta$ is set to 0.1. Training is done with Tensorflow \cite{abadi2016tensorflow} on a desktop with Intel i5-6500 CPU and a single NVIDIA Geforce GTX 1050 Ti GPU.\par

\begin{table}
\caption{The GWR-B hyperparameters we used in our experiments.}
\centering
\begin{tabular}{l c} 
 \toprule
 Hyperparameter & Value\\
 \midrule
 Activity Threshold & $a_T=0.8$\\
 Habituation Threshold & $h_T=0.15$\\
 Learning Rates & $\epsilon_c=0.1, \epsilon_n=0.01$\\
 Initial Habituation & $h_0=1$\\
 Habituation Curve & $\alpha_c=\alpha_n=1.05, \tau_c=3.3, \tau_n=14.3$\\
 Maximum Age Threshold & $\kappa=80$\\
\bottomrule
\end{tabular}
\label{table:table1}
\end{table}

\begin{figure}[t]
	\centering
		\includegraphics[width=0.65\linewidth]{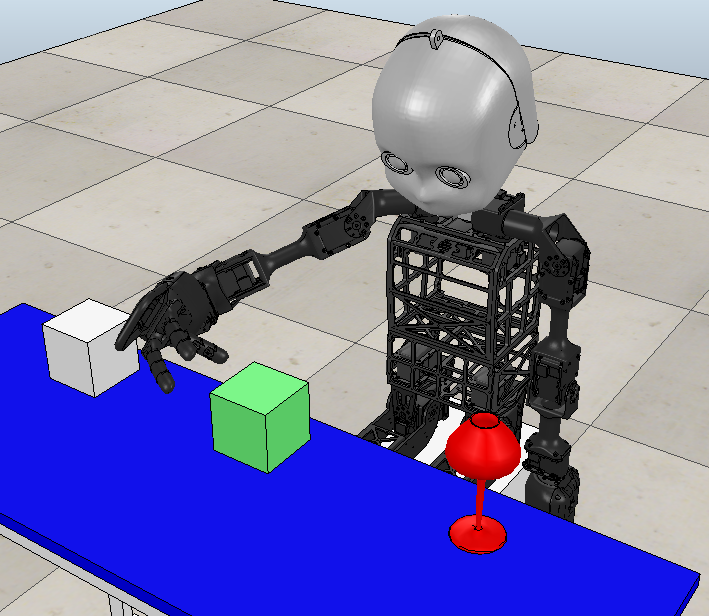}
		\caption{The CoppeliaSim simulation environment showing NICO facing a table with three objects.}
		\label{fig:4}
\end{figure}

\begin{figure}[t]
	\centering
		\includegraphics[width=\linewidth]{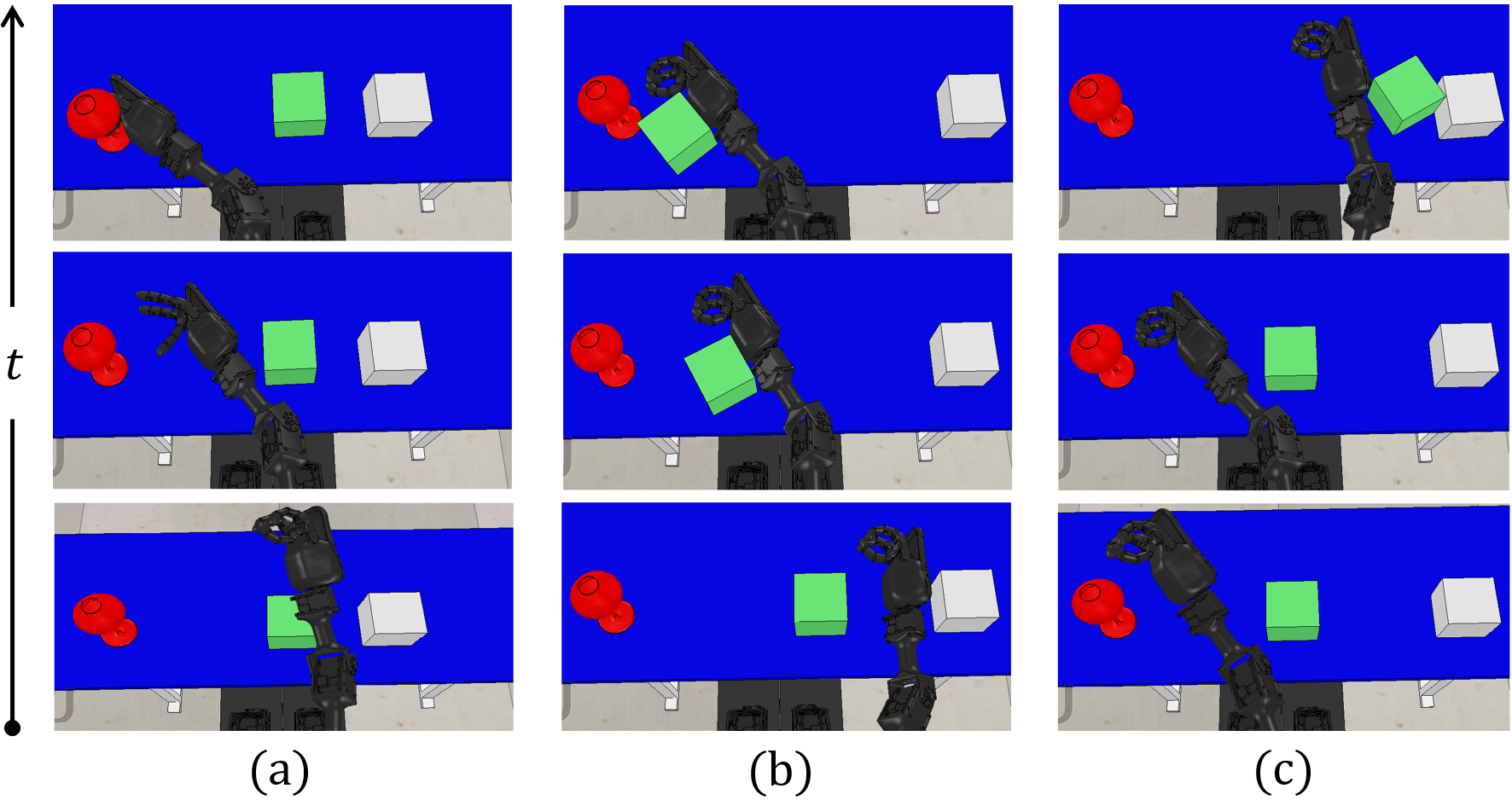}
		\caption{First-person demonstrations of three visuomotor tasks: (a) ``grasping the red glass", (b) ``pushing the green box towards the red glass", and (c) ``pushing the green box towards the white box". From bottom to top: RGB frames of initial, intermediate, and terminal configurations.}
		\label{fig:5}
\end{figure}

\noindent\textbf{Robot experiment setup:} All experiments are conducted on the Neuro-Inspired COmpanion (NICO) robot \cite{kerzel2017nico} using the CoppeliaSim (formerly V-REP) robot simulator \cite{rohmer2013v}. Fig. \ref{fig:4} shows the simulated NICO in front of a table on top of which three objects are placed. In all experiments, we consider a motor action involving four degrees of freedom in the right arm: two shoulder joints, one elbow joint, and one hand joint. The shoulder and elbow joints have an angular range of motion of $\pm{100}$ and $\pm{85}$ degrees respectively. The tendon-operated multi-fingered hand consists of 1 thumb and 2 index fingers with finger joints having an angular range of motion of $0{-}160$ degrees. The input to the learning system is a 64$\times$32 RGB image obtained from a vision sensor. Examples of the sensor's original output are shown in Fig. \ref{fig:5}.

\subsection{Multi-Task Learning Evaluation}
We consider the following visuomotor tasks: ``grasping the red glass" (Task 1), ``pushing the green box towards the red glass" (Task 2), and ``pushing the green box towards the white box" (Task 3). We collect 1200 first-person demonstrations per task with random initial configurations and object positions (see Fig. \ref{fig:5}). The demonstrations are, on average, about 30 steps ($\approx 6$s) long and are used for training the VAE and LAE models. Demonstration sequences are zero-padded to match a maximum sequence length of 50, and the timesteps with all-zero feature vectors are skipped using a masking layer when training the LAE model. Episodes terminate when the target object is grasped (Task 1), reached (Tasks 2 and 3), or after a maximum of 50 timesteps. A task is randomly sampled at the start of each episode, and the robot is given a sparse reward of +1 each time it completes the task. At the end of each episode, the BGPO algorithm is evaluated by randomly sampling a task and running the learned policy, given an input demonstration, for 50 timesteps. We compare the performance of BGPO to the MAML and Dist-NLL algorithms. The policy in MAML and Dist-NLL receives only the environment state as input. For implementing MAML, three replay buffers are used, one for each task. The policy that optimizes the performance over tasks after one gradient step on each task is tested on a randomly sampled task by collecting 256 adaptation trajectories and evaluating the policy finetuned on them for one gradient step in a test episode. Similarly, Dist-NLL is implemented using three task-specific policies, each having its own replay buffer, and a single multi-task policy with a separate output layer for each task. At each episode, the multi-task policy trains each output layer on 1k supervised examples of states sampled from a task-specific buffer and target actions given by the corresponding task-specific policy. The multi-task policy is then evaluated on a randomly sampled task in a test episode. We use DDPG as the base RL algorithm in MAML and Dist-NLL.\par
\begin{figure}[t]
	\centering
		\includegraphics[width=0.775\linewidth]{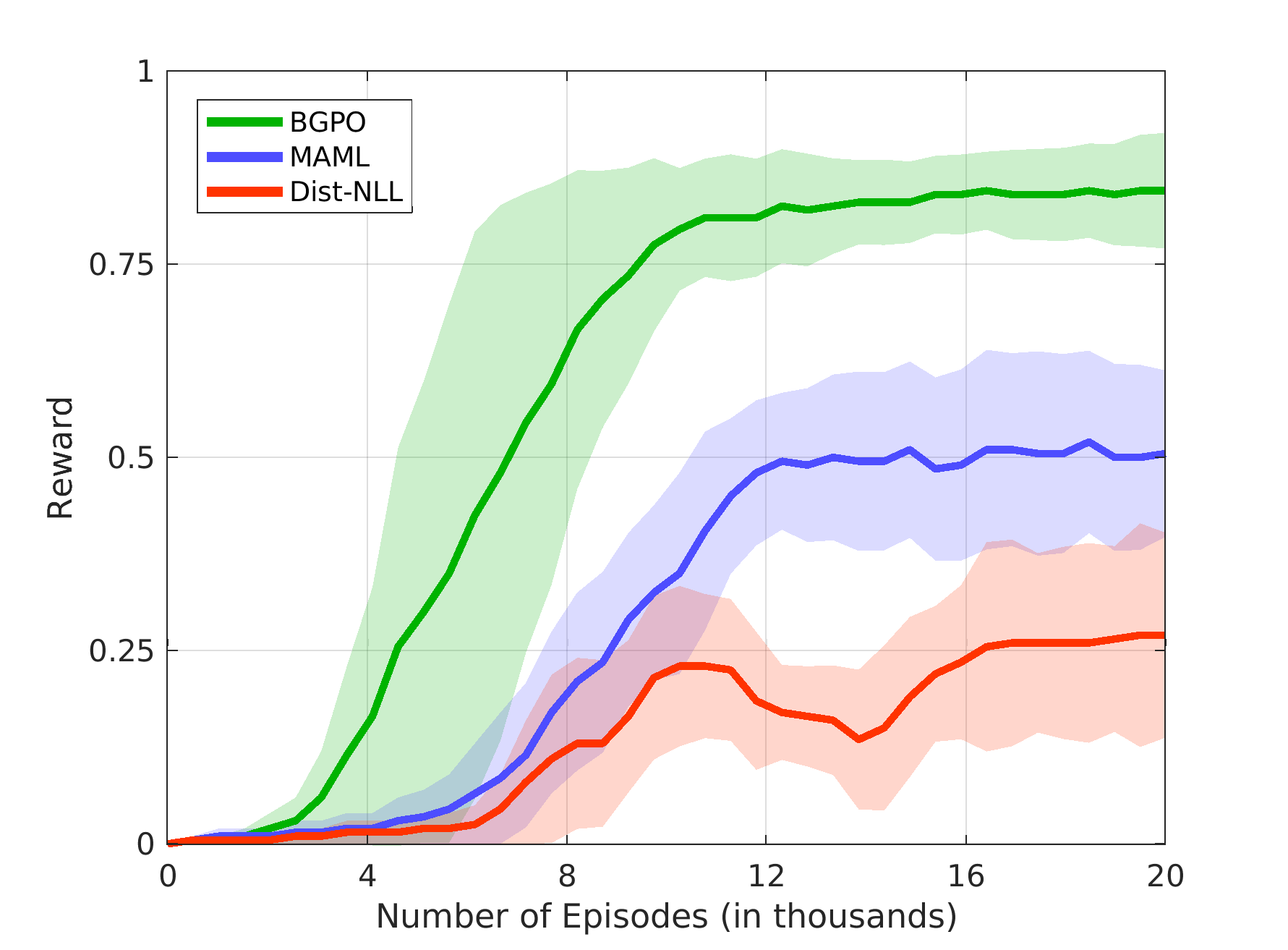}
		\caption{Performance curves of BGPO, MAML, and Dist-NLL on learning three independent visuomotor control tasks with the NICO robot. Shaded regions represent one standard deviation over 10 random seeds.}
		\label{fig:6}
\end{figure}

 Fig. \ref{fig:6} shows the total extrinsic reward per test episode for each algorithm, averaged over 10 random seeds. BGPO attains higher and more stable average rewards than MAML and Dist-NLL. Since the policy in MAML is updated by following the sum of gradients from each task, its generalizability strongly depends on the distance between individual gradient directions. This can be observed in the slow generalization of the learned policy which converged to an average performance. In Dist-NLL, the multi-task policy is trained only on supervised data from the task-specific policies that are still improving and are initially not well trained, resulting in instability and divergence. In contrast to these algorithms, BGPO avoids task-conditional updates in the parameter space of an unconditional policy by inferring the task in behavior space, steadily improving the performance and reaching over 80\% success rate.\par In general, BGPO learns not only to perform well on all tasks but also to complete a given task more efficiently than the demonstrations through its ability to infer the intended behavior behind a demonstration rather than to copy the observed actions. It is worth noting that BGPO performs the desired task even when the current environment state and robot configuration are considerably different from those in the first timestep of the input demonstration, as illustrated in the accompanying video (\url{https://youtu.be/4B4GL-HPQjg}).

\begin{figure}[t]
	\centering
		\includegraphics[width=0.8\linewidth]{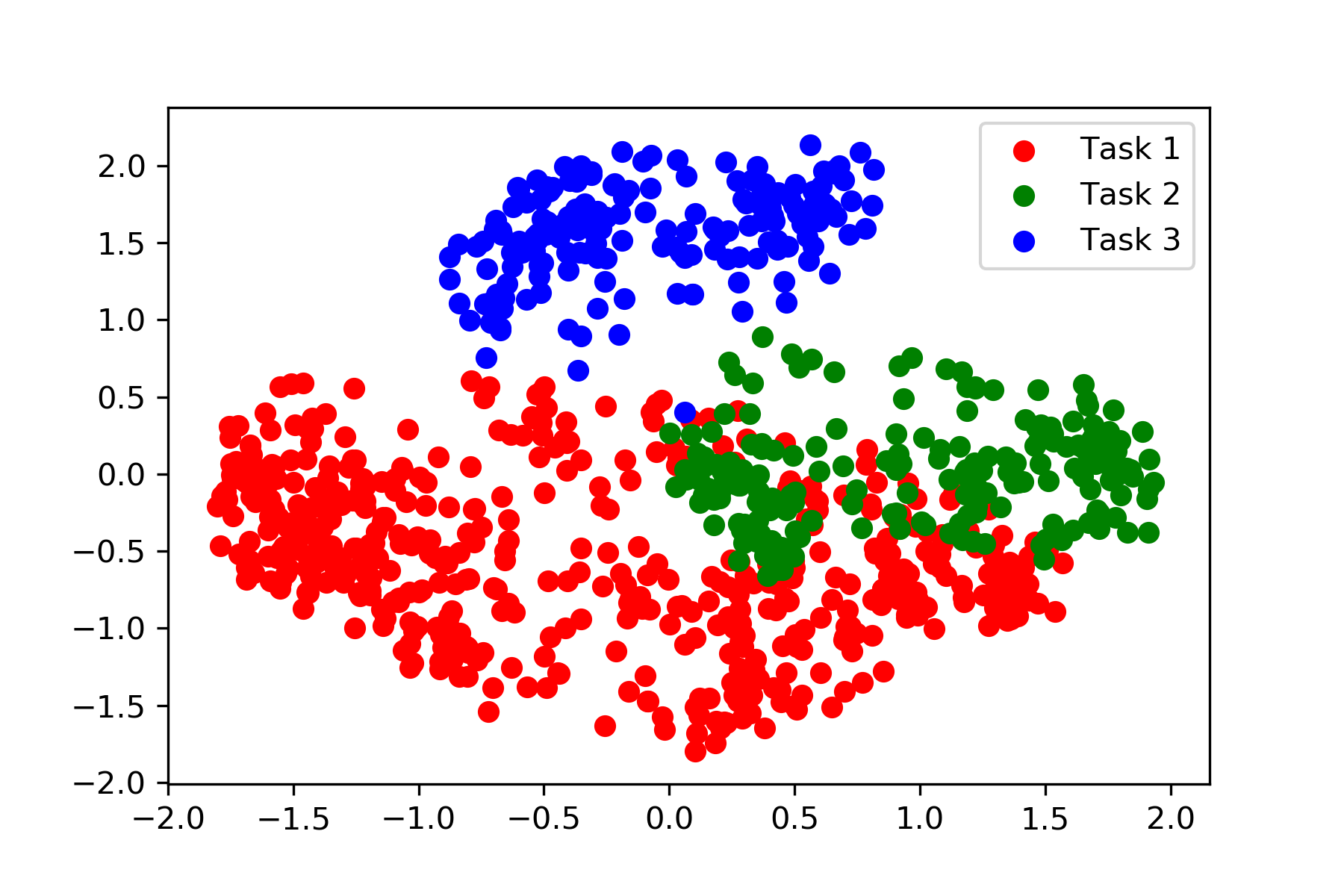}
		\caption{PCA visualization of the GWR node weights after one complete run of BGPO. The node color corresponds to the task with the highest number of demonstrations for which the node is the best match.}
		\label{fig:7}
\end{figure}

The trained GWR network has an average size of around 500 nodes over the 10 seeds. In order to visualize the GWR network at the end of learning, we use principal component analysis (PCA) to project the weight vectors onto a 2D space, as shown in Fig. \ref{fig:7}. The weights of the nodes that best match Task 1 or Task 2 demonstrations are clustered close together, reflecting the fact that both tasks involve a movement directed towards a shared object (i.e., the glass).

\subsection{Continual Learning Evaluation}
\begin{figure}[t]
	\centering
		\includegraphics[width=0.8\linewidth]{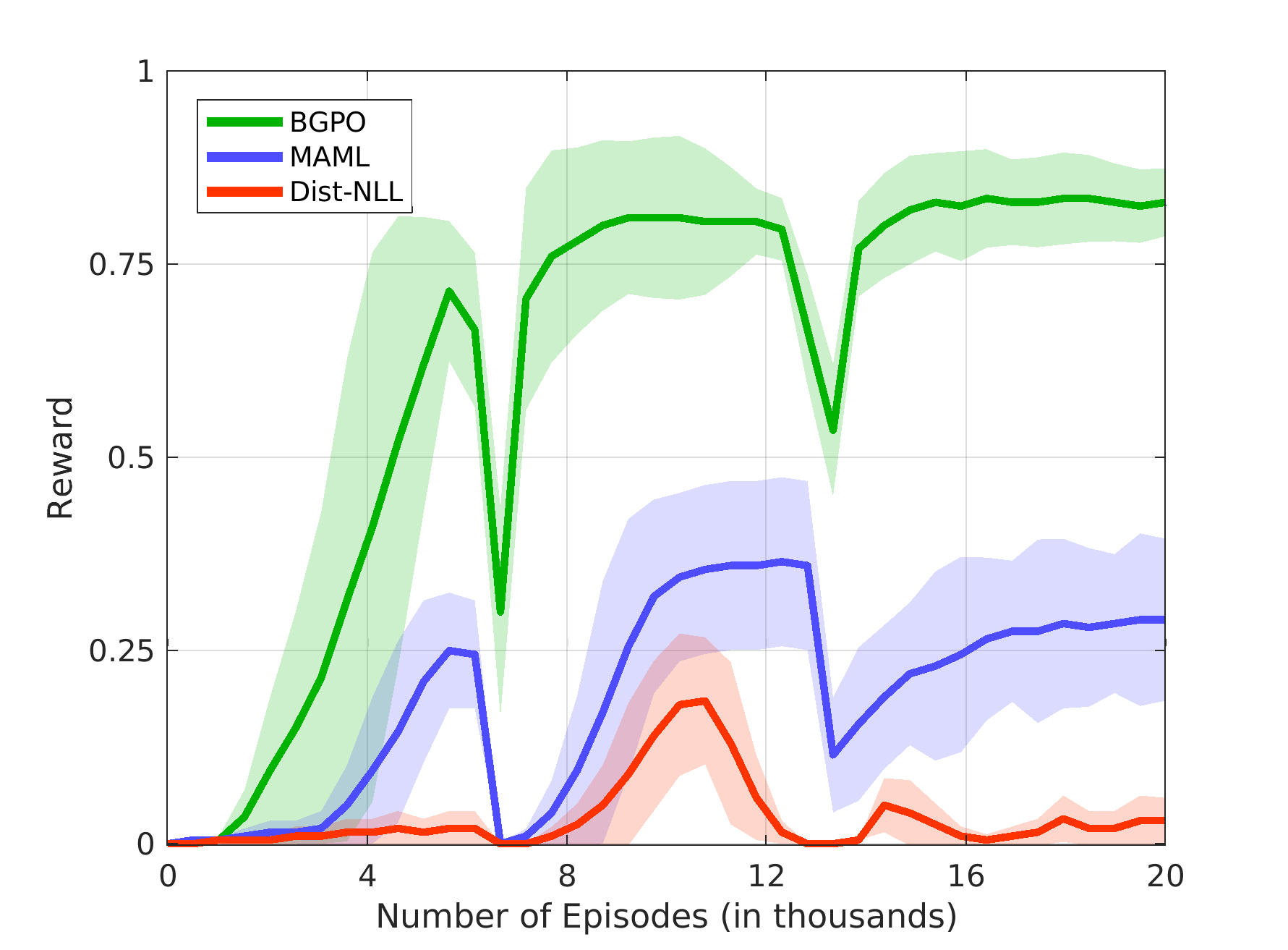}
		\caption{Performance curves of BGPO, MAML, and Dist-NLL in the continual learning setting. Shaded regions represent one standard deviation over 10 random seeds.}
		\label{fig:8}
\end{figure}

In the second experiment, we evaluate the performance of our algorithm and the baselines in a continual learning setting, where control tasks are presented sequentially. The learning starts with Task 1. Tasks 2 and 3 are presented after 7k and 13k episodes respectively. The learned policy is tested at the end of each episode for 50 steps on a task randomly sampled from the presented tasks. The results averaged over 10 random seeds are shown in Fig. \ref{fig:8}. In MAML, a new task causes a large training error that requires more data from the newly presented task to correct the direction of the sum of gradients over all tasks, resulting in slow convergence. This empirically shows why MAML requires the task distribution to be known a priori. While a predetermined number of tasks is necessary for Dist-NLL to know how many output layers the multi-task policy should have, we add a new output layer each time a new task is presented. Besides the non-stationary target actions from the constantly updated task-specific policies, adding output layers over time alters the learned parameters of the shared layers in the multi-task policy network and hence creates further learning divergence. BGPO, on the other hand, is specially devised to cope with sequentially introduced tasks through the ability of its self-organizing network to grow when it does not have a close match to a demonstrated behavior. This adaptation occurs in behavior space without disturbing the policy parameters or architecture, allowing faster task adaptation.

\subsection{Ablation Study}
\begin{figure}[t]
	\centering
		\includegraphics[width=0.8\linewidth]{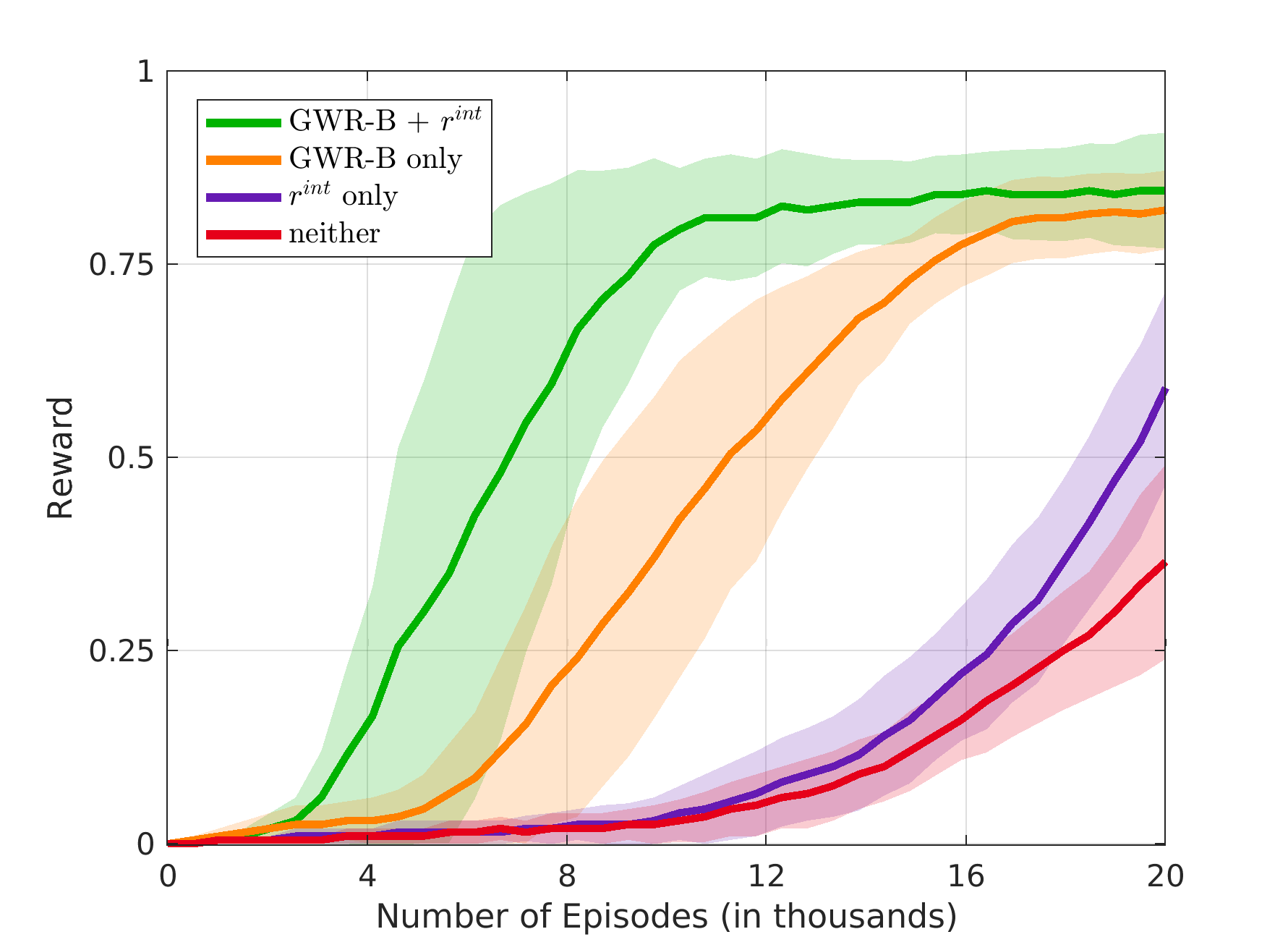}
		\caption{Performance curves of BGPO with its two components (GWR-B$+r^{int}$), with the GWR-B model only, with the intrinsic reward only, and with both components removed. Shaded regions represent one standard deviation over 10 random seeds.}
		\label{fig:9}
\end{figure}

To analyze the contribution of each component of the BGPO algorithm, namely the GWR-B model and the behavior-matching intrinsic reward, we conduct an experiment where we compare the performance of BGPO after removing each component in turn and in combination. Fig. \ref{fig:9} shows the results for the test episodes averaged over 10 random seeds. When the GWR-B model is removed, the algorithm takes the encoded demonstration from the LAE encoder as input instead of the behavior embedding and optimizes the policy over the joint state-demonstration space. Since the encoded demonstration by itself captures no information about the behavior space, the algorithm lacks the ability to infer the intended behavior behind the demonstration, thereby slowing down the convergence of the policy. Adding only the behavior-matching intrinsic reward, where the behavior embedding is replaced with the encoded demonstration, slightly improves the performance by constantly providing informative feedback in the absence of an extrinsic reward, as shown in Fig. \ref{fig:9}. However, we observe that adding only the GWR-B model significantly improves the performance, since using the behavior embedding from the model allows to generalize the policy learned for this embedding to input demonstrations for which the embedding is the best match. Combining the GWR-B model with the behavior-matching intrinsic reward leads to faster convergence, compared to the GWR-B model alone, by incentivizing the policy to generate trajectories close in behavior space to the behavior embedding.

\section{Conclusion}
This paper introduced a novel approach to continual robot learning of visuomotor tasks. Our approach solves task inference by learning a behavior embedding space from incremental self-organization of behavior demonstrations. To optimize performance over tasks, our approach learns a policy over the joint state-behavior space. The distance in behavior space between demonstrated and generated trajectories is used to derive a behavior-matching intrinsic reward, encouraging the policy to follow the intended behavior behind the demonstration. In contrast to previous approaches, our approach makes no assumptions about task distribution or policy architecture and avoids disruptive task-conditional updates to an unconditional policy. The results show that our approach achieves better generalization performance and convergence speed compared to multi-task learning baselines, particularly in the continual learning setting. 

The tasks considered in the present work are structurally independent of each other. Extending the proposed approach to interdependent tasks can allow for continual learning of more complex hierarchical skills by incorporating techniques such as multi-level control and temporal abstraction of actions \cite{sutton1999between}. Transforming third-person demonstrations into first-person demonstrations in our approach, for example, by training a context translation model \cite{liu2018imitation}, is an exciting avenue for future work that would enable continual learning from a human teacher in social human-robot interaction.

\section*{Acknowledgement}
We gratefully acknowledge support from the German Research Foundation DFG under project CML (TRR 169).

\bibliographystyle{IEEEtran}
\bibliography{bibliography}
\end{document}